\documentclass{article}

\usepackage{arxiv}

\usepackage[utf8]{inputenc} 
\usepackage[T1]{fontenc}    
\usepackage{hyperref}       
\usepackage{url}            
\usepackage{booktabs}       
\usepackage{amsfonts}       
\usepackage{nicefrac}       
\usepackage{microtype}      
\usepackage{lipsum}
\usepackage{graphicx}

\usepackage{multirow}
\usepackage{pifont}
\usepackage{makecell}

\graphicspath{ {./images/} }

\title{Multimodal Conditional MeshGAN for Personalized Aneurysm Growth Prediction}

\author{
 Long Chen \\
  MRC Laboratory of Medical Sciences\\
  Imperial College London\\
   \And
 Ashiv Patel \\
  Imperial College Healthcare\\
  NHS Trust\\
  \And
 Mengyun Qiao \\
  Department of Mechanical Engineering\\
  University College London\\
  \And
 Mohammad Yousuf Salmasi \\
  Imperial College Healthcare\\
  NHS Trust\\
  \And
 Salah A. Hammouche \\
  London Postgraduate School of Surgery\\
  NHS England\\
  \And
 Vasilis Stavrinides \\
  Imperial College Healthcare\\ 
  NHS Trust\\
  \And
 Jasleen Nagi \\
  Faculty of Medicine\\
  Imperial College London\\
  \And
 Soodeh Kalaie \\
  MRC Laboratory of Medical Sciences\\
  Imperial College London\\
  \And
 Xiao Yun Xu \\
  Department of Chemical Engineering\\
  Imperial College London\\
  \And
 Wenjia Bai \\
  Department of Brain Sciences\&Computing\\
  Imperial College London\\
  \And
 Declan P. O'Regan$^{*}$ \\
  MRC Laboratory of Medical Sciences\\
  Imperial College London\\
}

\begin{document}
\maketitle
\begin{abstract}
Personalized, accurate prediction of aortic aneurysm progression is essential for timely intervention but remains challenging due to the need to model both subtle local deformations and global anatomical changes within complex 3D geometries. We propose MCMeshGAN, the first multimodal conditional mesh-to-mesh generative adversarial network for 3D aneurysm growth prediction. MCMeshGAN introduces a dual-branch architecture combining a novel local KNN-based convolutional network (KCN) to preserve fine-grained geometric details and a global graph convolutional network (GCN) to capture long-range structural context, overcoming the over-smoothing limitations of deep GCNs. A dedicated condition branch encodes clinical attributes (age, sex) and the target time interval to generate anatomically plausible, temporally controlled predictions, enabling retrospective and prospective modeling. We curated TAAMesh, a new longitudinal thoracic aortic aneurysm mesh dataset consisting of 590 multimodal records (CT scans, 3D meshes, and clinical data) from 208 patients. Extensive experiments demonstrate that MCMeshGAN consistently outperforms state-of-the-art baselines in both geometric accuracy and clinically important diameter estimation. This framework offers a robust step toward clinically deployable, personalized 3D disease trajectory modeling. The source code for MCMeshGAN and the baseline methods is publicly available at \url{https://github.com/ImperialCollegeLondon/MCMeshGAN}.
\end{abstract}


\section{Introduction}
\label{sec:introduction}
Personalized prediction of disease trajectories is a cornerstone of precision medicine, facilitating diagnosis and intervention at an appropriate stage. In the context of aortic aneurysm imaging, predictive anatomical modeling techniques based on medical images have garnered increasing attention. Aortic aneurysms are irreversible, localized dilatations of the aorta that, if left untreated, can lead to rupture and life-threatening outcomes  \cite{chung2024artificial}. Therefore, accurate prediction of aneurysm growth is critical for effective surveillance and treatment planning. Conventional approaches typically rely on manual measurements of aneurysm diameter from follow-up imaging \cite{anfinogenova2022existing}, coupled with empirical diameter thresholds to guide clinical decisions on surgical intervention \cite{elefteriades2023ascending}. However, these methods are primarily limited to 2D representations, which are insufficient for capturing the complex 3D morphological changes involved in aneurysm progression. While existing studies focus on 2D image-based prediction, direct modeling on 3D anatomical surfaces, such as meshes, remains relatively underexplored and represents a promising direction for advancing anatomical growth modeling.

Specifically, prior efforts in aneurysm growth prediction can be broadly categorized into biomechanically inspired models and conventional machine learning methods. Biomechanical models, such as growth and remodeling (G$\&$R) frameworks \cite{zhang2019patient}, are grounded in physiological principles but rely heavily on predefined material properties and assumptions. While these models offer interpretability, they are often computationally expensive and difficult to generalize across diverse patient populations. In contrast, machine learning approaches such as Gaussian process models \cite{do2018prediction}, support vector machines (SVMs) \cite{geronzi2023computer}, and deep belief networks (DBN) \cite{jiang2020deep} depend on handcrafted features, which may fall short in capturing the complex 3D geometric changes associated with aneurysm progression. These models often perform well on small datasets but struggle to scale to unseen clinical cases due to their limited capacity to learn rich spatial representations. Meanwhile, deep learning techniques like Convolutional Neural Networks (CNNs) are well-suited for processing Euclidean grid-structured data, but they are inherently incompatible with irregular, non-Euclidean structures likes 3D meshes. To address this, Graph Convolutional Networks (GCNs) \cite{kipf2017semi, ranjan2018generating, zhang2023meshwgan} have been introduced to work directly on graph-based 3D data such as meshes and point clouds. While GCNs provide a natural way to handle mesh topology, they still struggle to simultaneously capture both global context and preserve local geometric details. In particular, stacking multiple GCN layers to expand the receptive field often leads to over-smoothing effects \cite{min2020scattering, rusch2023survey}, which reduces the network’s ability to maintain fine-grained local fidelity and makes it less effective at modeling the subtle yet clinically important shape changes involved in aneurysm progression.

To address the limitations of existing methods, we propose MCMeshGAN, a multimodal, conditional 3D mesh-to-mesh prediction framework for personalized aneurysm growth modeling. Most previous studies rely on 2D image-based predictions \cite{zhang2019patient, geronzi2023computer, jiang2020deep}, which are limited in capturing the complex 3D shape changes of aneurysms. In contrast, MCMeshGAN directly operates on 3D surface meshes, enabling more anatomically faithful modeling of geometric changes over time. To mitigate loss of local geometric details by stacking GCNs, MCMeshGAN introduces a dual-branch architecture that combines a global GCN branch to encode the overall anatomical structure with a local KNN-based convolutional branch that preserves fine-grained surface details and improves local deformation prediction. This design effectively balances global structure and local detail, overcoming key limitations of previous GCN-only models. In addition, earlier works typically ignore clinical metadata, which is crucial for personalized modeling. MCMeshGAN incorporates multimodal learning by integrating patient-specific information (e.g., age, sex, and time interval) through a dedicated condition branch, enabling personalized and temporally adaptive predictions. In summary, our key contributions are as follows:
\begin{itemize}
    \item To the best of our knowledge, MCMeshGAN is the first multimodal conditional mesh-to-mesh translation model for 3D aneurysm growth prediction. It enables continuous, conditional mesh generation in a controlled and clinical information-guided manner.
    \item MCMeshGAN incorporates a global graph branch for modeling global context and a local KNN-based convolutional branch for extracting fine-grained local features. These two branches facilitate the generation of the overall 3D geometry and the prediction of subtle geometric changes in the local region, respectively.    
    \item MCMeshGAN integrates information from multiple modalities, including clinical text (e.g., age and sex) and 3D geometric data (meshes). Pairing clinical information with 3D geometry helps the model to better capture the relationship between a patient's aneurysm morphology and their clinical characteristics, resulting in more realistic and personalized predictions.
    \item We build a 3D mesh dataset that includes longitudinal mesh data from patients with thoracic aortic aneurysms, facilitating the advancement of mesh-to-mesh translation for aneurysm growth prediction.
\end{itemize}

\section{Related Works}
\label{sec:related}

\subsection{3D Transformation Models}
3D transformation models are extensively used in computer graphics and geometric data processing such as mesh-to-mesh and point cloud-to-point cloud translations \cite{zhao2023multi, zhang2023meshwgan, lee2022pu}. This process transforms one 3D data representation into another, typically altering features such as shape and structure, while preserving essential geometric or semantic properties. 3D transformation models have been applied in various tasks, including point cloud generation \cite{luo2021diffusion}, mesh denoising \cite{wang2024hyper}, and facial expression modeling \cite{olivier2023facetunegan}. 

Some early studies leverage CNNs, such as SpiralNet++ \cite{gong2019spiralnet++}, for 3D graph-structured data processing. CNNs have proven effective for processing data on a regular grid in the Euclidean space, but adapting them to data with irregular structures presents challenges \cite{Bronstein2017}. This is because CNNs require regular input, limiting their application to irregular structures like meshes and point clouds. To overcome this, various geometric representations have been developed to convert irregular mesh data into regular structures \cite{wang2019data, wei2019mesh}. However, CNNs are inherently constrained by their local convolution operations, which lack the ability to capture global context, thus restricting their generalization to complex scenes. 

In contrast, GCNs \cite{kipf2017semi, ranjan2018generating} may be better suited to processing graph-structured data, which naturally exhibit a graph structure defined by vertices and face connectivity. \cite{chen2020simple} introduced GCN2Conv, an extension of the vanilla GCN that incorporates residual and identity mappings to mitigate the over-smoothing issue, where GCN performance deteriorates as depth increases. \cite{zhu2021simple} proposed simple spectral graph convolution (SSGConv), derived from a modified Markov diffusion kernel, which captures both global and local contexts of each node in the graph. \cite{yang2022graph} introduced PMLP, an intermediate model class that bridges MLP and GCN by utilizing an MLP architecture in training and a GCN architecture during testing. PMLPs consistently perform on par with, or even surpass, their GCN counterparts. Although GCNs are highly effective for processing graph-structured data, they often face challenges in balancing local detail with global structural context. To capture broader context, GCNs typically stack multiple layers. However, this strategy can lead to the well-known over-smoothing issue \cite{min2020scattering}. As network depth increases, node feature embeddings lose their individual characteristics, which blurs local distinctions, especially among neighboring nodes \cite{rusch2023survey}. This not only diminishes the model's ability to distinguish between different nodes but also lead to the loss of fine-grained local details, resulting in unrealistic local geometric representations. Consequently, GCNs may fail to capture subtle local variations that are critical for accurately modeling complex phenomena such as disease progression. 

\subsection{Aneurysm Growth Prediction}
\normalcolor

Most research on predicting aneurysm growth depends on manual measurements of the aortic diameter through imaging techniques \cite{anfinogenova2022existing}. For instance, \cite{gharahi2015growth} utilized the maximally inscribed sphere (MIS) method to extract the diameter from 3D images.  Previous efforts to predict aortic aneurysm progression have explored biomechanically-inspired models and data-driven machine learning models. For example, \cite{zhang2019patient} used a G$\&$R model to forecast the future expansion of an abdominal aortic aneurysm (AAA), while \cite{do2018prediction} developed a dynamical Gaussian process implicit surface (DGPIS) model to predict AAA growth from longitudinal CT scans. Meanwhile, \cite{geronzi2023computer} employed handcrafted local and global shape features alongside traditional machine learning techniques, such as SVM and regression models, to forecast the growth of ascending aortic aneurysms. In contrast, \cite{jiang2020deep} trained a two-layer DBN on both real and simulated data to enable fast, personalized predictions of AAA expansion. Despite their effectiveness, these approaches are limited by their dependence on biomechanical priors or handcrafted features, which restricts their ability to generalize.

In recent years, deep learning has gained significant attention, though few studies have explored the use of deep neural networks for aneurysm growth prediction. The pioneering study by \cite{kim2022deep} developed a 2D patch-based CNN, manually selecting multiphysical features as input to predict the growth of AAA. However, this approach focuses on 2D patch data, which fails to capture the 3D geometrical changes occurring during aneurysm progression. In contrast, our work aims to predict patient-specific aneurysm evolution using a 3D mesh-to-mesh translation network.

\begin{figure*}[htp]
  \centering
   \includegraphics[width=1.0\linewidth]{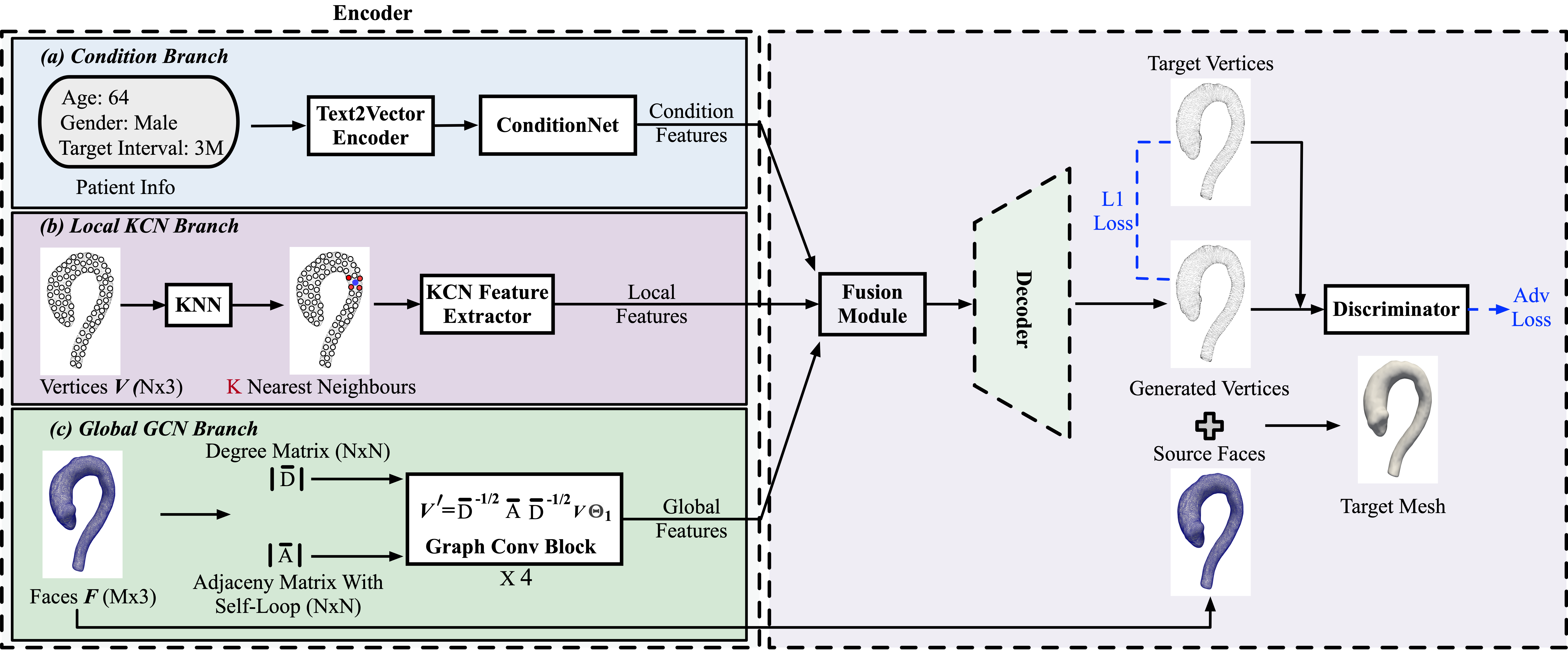}
   \caption{Overview of the MCMeshGAN architecture: (a) The condition branch receives clinically relevant information, with the target time interval serving as a key factor that guides the predicted aneurysm progression. (b) The KCN branch leverages the KNN algorithm to capture fine-grained variations within local neighborhoods, while (c) the GCN branch models the global anatomical structure of the mesh. The fusion module integrates features extracted from multiple branches, facilitating the capture of both fine details and broader structural patterns within the mesh.}
   \label{fig:ageaortagan}
\end{figure*}

\section{Method}
\label{sec:formatting}

We propose MCMeshGAN, a multimodal conditional mesh-to-mesh translation network designed to predict 3D geometric changes in an aneurysm during its growth. In this section, we first provide an overview of MCMeshGAN, followed by a detailed description of each component.

\subsection{Method Overview: MCMeshGAN}

Fig. \ref{fig:ageaortagan} depicts the proposed MCMeshGAN model, which comprises three branches: a condition branch for extracting clinical information as the condition features, a local KNN-based convolutional (KCN) branch that employs the K nearest neighbors (KNN) algorithm to capture local geometric features of the vertices on the input mesh, and a global graph convolutional (GCN) branch that leverages the graph structure to extract global structural features and contextual information from both vertices and edges within the mesh. 

As human aortic aneurysms exhibit significant variability in geometry and shape due to factors such as age, sex, and ethnicity \cite{tang2016lifetime}, MCMeshGAN incorporates a condition branch that leverages patient-specific clinical information, including age, sex, and the target time interval, to support personalized and accurate predictions. Furthermore, accurate aneurysm growth modeling requires understanding both global anatomical changes and subtle local deformations, particularly those that are clinically significant. To capture this multiscale information, MCMeshGAN employs two complementary branches: a KCN branch and a GCN branch. The GCN branch stacks multiple graph convolutional layers to expand the receptive field and model the global anatomical structure. However, adding more GCN layers often leads to over-smoothing, causing the loss of fine local geometric details. To overcome this limitation, the KCN branch uses $k$-nearest neighbor-based convolution to extract fine-grained local features from each vertex and its surrounding neighborhood. This dual-branch design allows MCMeshGAN to effectively balance global structural understanding with fine-grained local detail.

\begin{figure}[ht]
  \centering
   \includegraphics[width=0.7\linewidth]{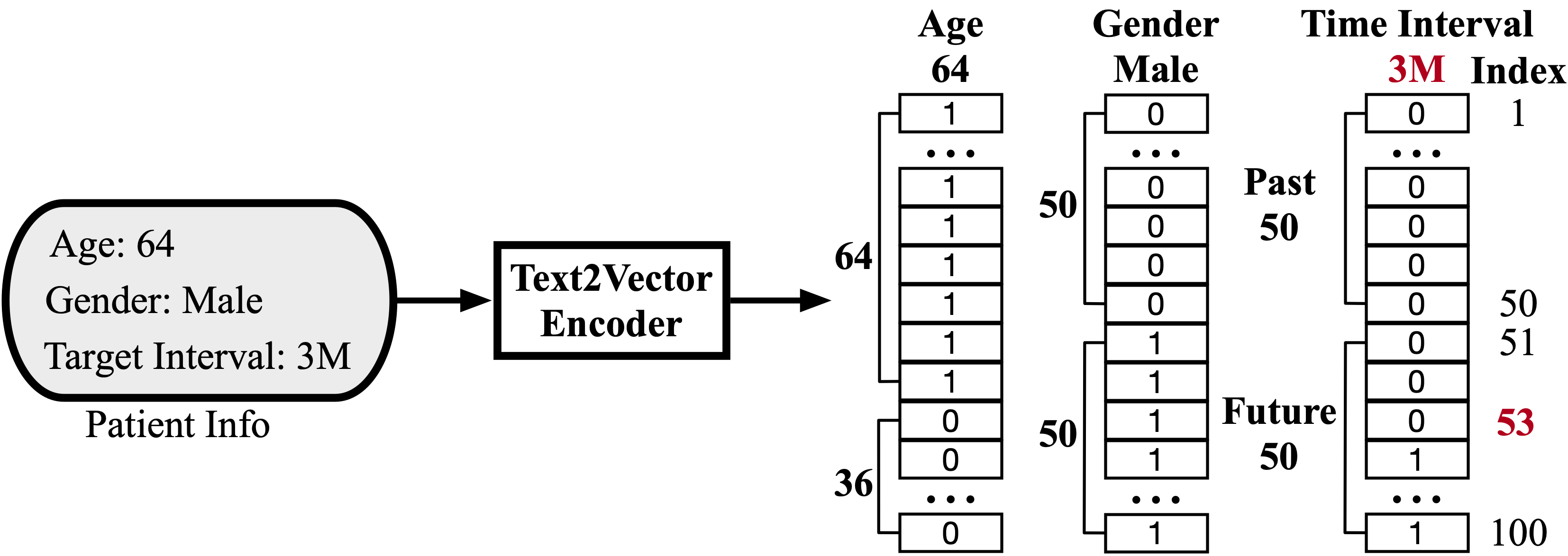}
   \caption{Text2Vector transforms age, sex, and target interval into vectors using ordinal binary encoding. The time\_interval\_vector $\in \mathbb{R}^{1\times 100}$ encodes the time interval (e.g., time interval=$+3$), which can be either positive (for future prediction) or negative (for past reconstruction). Its values are defined as: time\_interval\_vector$[1:50+$time interval$]=0$ and time\_interval\_vector$[50+$time interval$+1:100]=1$}
   \label{fig:text2vector}
\end{figure}
\subsection{The Condition Branch}
The condition branch enables MCMeshGAN to generate the aneurysm meshes over time, explicitly controlled by the input target time interval and personalized clinical information. Given that aneurysm geometry and shape vary with factors such as age and sex, we incorporate these attributes as additional inputs. At the training stage, given a source aneurysm mesh data $m_s$ with clinical information (source age: $a_s$, sex: $g_s$, and data collection time $t_s$), we aim to predict a target mesh of the same patient $m_t$ at a different time point with its clinical information (age: $a_t$, sex: $g_t$, and data collection time $t_t$). The target time interval is then calculated as $time\_{interval}=t_t-t_s$. At the inference stage, given a source aneurysm mesh data $m_s$ along with age $a_s$ and sex $g_s$ information, we can specify a target time interval $time\_{interval}$ and MCMeshGAN can predict the target aneurysm mesh $m_t =G(m_s, a_s, g_s|time\_{interval})$, conditioned on the specified time interval. 

To incorporate clinical information into our model, we design a Text2Vector encoder that uses the ordinal binary vectors to encode age, sex, and time interval. Each clinical attribute is represented as a 100 $\times$ 1 binary vector, as illustrated in Fig. \ref{fig:text2vector}. The time interval represents the temporal difference between the baseline and target time points (e.g., a time interval of 3 months in Fig. \ref{fig:text2vector}). This value can be either positive (e.g., $+$3, indicating prediction of the aneurysm shape 3 months after the baseline scan) or negative (e.g., $-$3, indicating reconstruction of the shape 3 months before the baseline scan). To encode this temporal information, we define an ordinal binary vector time\_interval\_vector $\in \mathbb{R}^{1\times 100}$, with values set as: time\_interval\_vector$[1:50+$time interval$]=0$ and time\_interval\_vector$[50+$time interval$+1:100]=1$. Here, index 50 serves as the dividing point between past and future.

\begin{figure*}[h]
  \centering
   \includegraphics[width=1.0\linewidth]{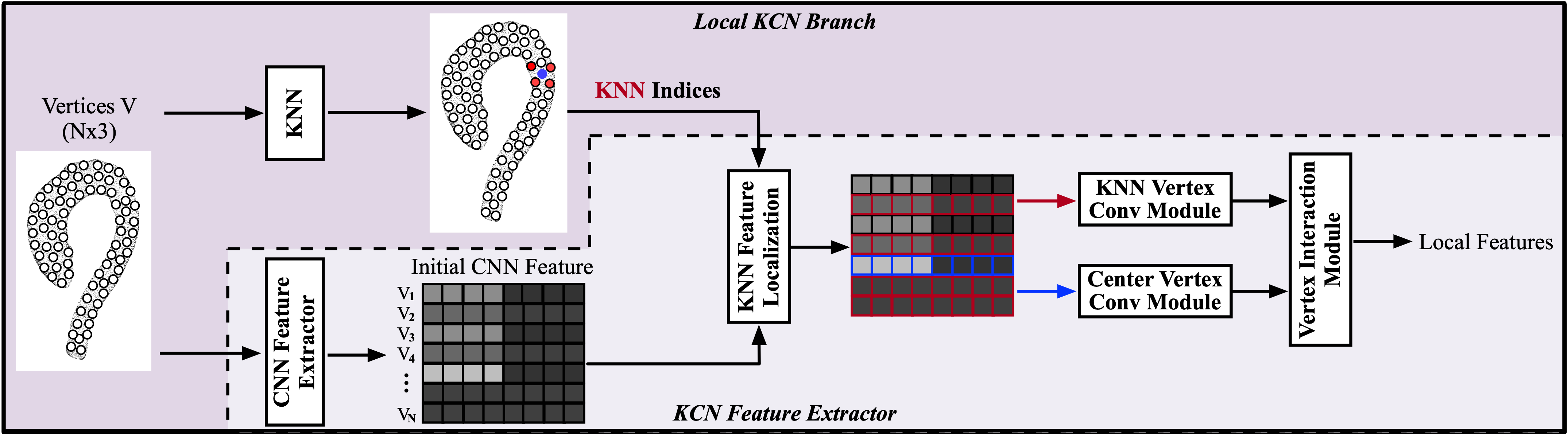}
   \caption{Overview of the KCN branch. The vertex coordinates are represented as a 2D data array and a CNN feature extractor is applied to extract the CNN features. For each vertex, the KNN algorithm is used to identify its K nearest neighbors within the CNN feature space. Two parallel convolutional modules, the KNN Vertex Convolution module and the Center Vertex Convolution module, are used to process features from each vertex and its local $k$-nearest neighbors. Finally, a vertex interaction module models the interactions between each vertex and its neighbors.}
   \label{fig:KCN}
\end{figure*}
\subsection{The Local KCN Branch}
To adapt CNNs for mesh data, we introduce a KCN branch that integrates local convolutional feature extraction with a KNN-based sampling strategy, preserving detailed features at local vertex level. Fig. \ref{fig:KCN} presents the details of the KCN branch. Let $V\in \mathbb{R}^{N\times d}$ denote the set of mesh vertices, where $N$ is the number of vertices and each vertex $\mathbf{ v_i} \in \mathbb{R}^d$ is represented by its 3D coordinates. We first construct a CNN-based feature extractor $\mathcal F_{CNN}$ to transform the raw 3D coordinates of each vertex into an informative per-vertex feature representation $\mathcal F(V)$:
\begin{equation}
  \mathcal F(V)=\mathcal F_{CNN}(V)
  \label{eq:barD}
\end{equation}
where $\mathcal F(V)\in \mathbb{R}^{N\times d_f}$, and $d_f$ is the feature dimension after the CNN transformation. Then, for each query vertex $q \in V$, we apply the KNN algorithm to find its $K$ nearest neighbors $\mathcal{N}(q)$, and use the indices of these neighbors to localize their KNN features in the initial CNN feature.
\begin{equation}
  \mathcal{N}(q) = \{v_{i_1}, v_{i_2}, ..., v_{i_k}\} 
  \label{eq:barD}
\end{equation}
where $i_1$, $i_2$, ..., $i_k$ represent the indices of the $k$ nearest neighbors. These KNN indices are used to localize the KNN features $\mathcal{F}(\mathcal{N}(q))$ within the initial CNN feature map. 
\begin{equation}
  \mathcal{F}(\mathcal{N}(q)) = \{\mathcal{F}(v_{i_1}), \mathcal{F}(v_{i_2}), ..., \mathcal{F}(v_{i_k})\} 
  \label{eq:barD}
\end{equation}
Next, two parallel convolutional branches $\mathcal{F}_{Center}$ and $\mathcal{F}_{Nei}$ extract features from each vertex and its $k$ nearest neighbors, respectively. A vertex interaction module then captures the relationships between each vertex and its neighbors, producing the final local features. This module consists of a feature concatenation layer, $\mathcal F_C$, followed by a feature mean function $\mathcal F_M$. The resulting local features $\mathcal{F}_{local}$ can be represented as
\begin{equation}
  \mathcal{F}_{local} = \mathcal F_M (\mathcal F_C [\mathcal{F}_{Nei}(\mathcal F (\mathcal{N}(q))), \mathcal{F}_{Center}(F (q))] )
  \label{eq:barD}
\end{equation}
The local KCN branch explicitly gathers information on how vertices relate to their $k$ nearest neighbors, aiding in the capture of fine details in the vicinity of the vertices. As a result, it is sensitive to small-scale variations and changes in local regions. This ability enables it capture precise geometric changes around the aneurysm areas, which are crucial for subsequent measurements (such as the the maximum aneurysm diameter) and diagnosis.

\begin{figure*}[t]
  \centering
   \includegraphics[width=1\linewidth]{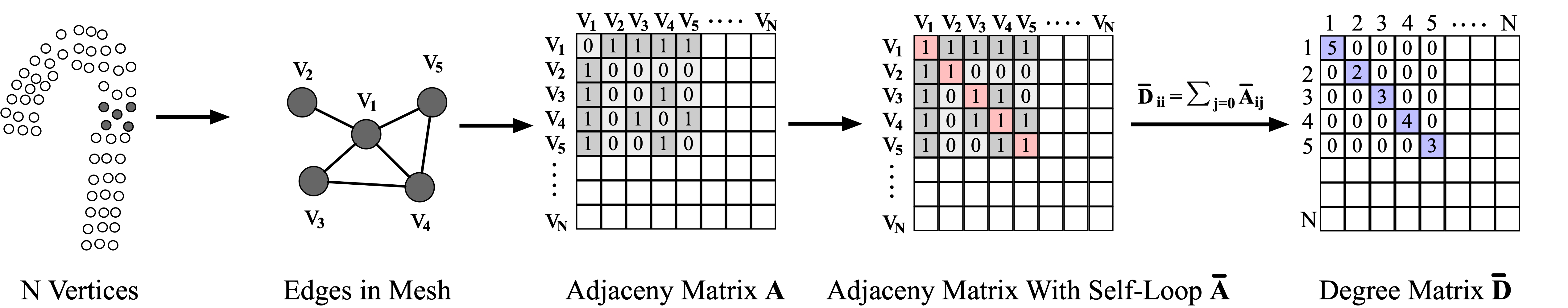}
   \caption{The adjacency matrix $A$, the self-looped adjacency matrix $\bar A$, and the degree matrix $\bar D$ in the graph branch are derived from the edges connecting vertices in the mesh.}
   \label{fig:GCN}
\end{figure*}
\subsection{The Global GCN Branch}
\normalcolor

A 3D mesh can be represented as an undirected graph $\mathcal G =(V, \xi)$, where $|V|=N$ denotes the number of vertices and $\xi$ represent the edges. The edge set $\xi$ $\in \mathbb{R}^{M \times 2}$ represents the graph's connectivity, for which a binary adjacency matrix $A \in \mathbb{R}^{N \times N}$ is often used to encode the edge connections. In this matrix, $A_{ij}=1$ indicates an edge connecting vertices $v_i$ and $v_j$, while $A_{ij}=0$ indicates no connection. The GCN branch is composed of four graph convolutional blocks. Each block is built using the graph convolutional layers described in \cite{kipf2017semi}, which extract features from both nodes and edges. For each graph convolutional layer, the nodes communicate with their neighbors through edges to exchange information. This message passing mechanism can be defined by the graph convolutional operation between nodes and edges, formulated in Eq. \ref{eq:GCN}:
\begin{equation}
  V' = \bar{D}^{-1/2}\bar{A}\bar{D}^{-1/2}V\Theta
  \label{eq:GCN}
\end{equation}
Here, $V$ and $V'$ represent the input and updated vertex coordinates respectively, while $\Theta$ denotes the parameters of the GCN layer. $\bar{A}$ and $\bar{D}$ are derived from the adjacency matrix $A$ of the graph. The matrix $\bar{A} \in \mathbb{R}^{N \times N}$ is the adjacency matrix of the undirected graph $\mathcal G$ with added self-connections, expressed as $\bar{A}= A+ I_N$, where $I_N$ is the identity matrix. $\bar{D} \in \mathbb{R}^{N \times N}$ represents the diagonal degree matrix of $\bar{A}$, which measures the degree of each node, and is defined as $\bar{D}_{ii} = \sum_j \bar{A}_{ij}$. An example of $\bar{A}$ and $\bar{D}$ is visualized in Fig. \ref{fig:GCN}.

The GCN branch captures global structural relationships by passing messages between nodes and across multiple layers. By stacking four graph convolutional layers, the GCN branch gathers information from increasingly distant nodes, allowing it to capture a broader context throughout the graph. This process is crucial for understanding the overall geometry and structure of the mesh.

\begin{figure}[ht]
  \centering
   \includegraphics[width=0.65\linewidth]{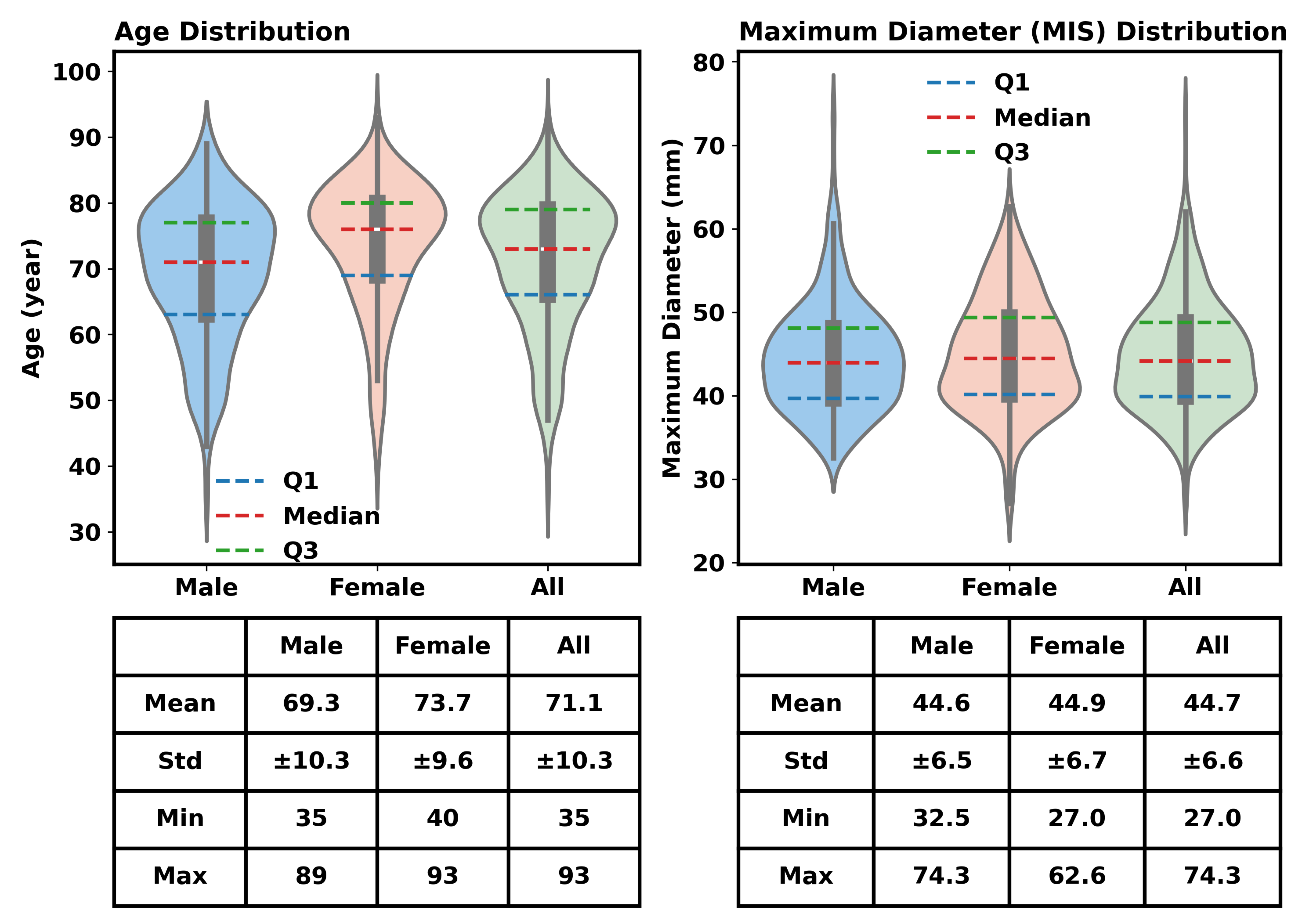}
   \caption{Summary statistics for patient age and maximum inscribed sphere (MIS) diameter in the TAAMesh dataset.}
   \label{fig:statistics}
\end{figure}

\section{Experimental Setup}
\subsection{Datasets and Evaluation Metrics}
Due to the lack of publicly available longitudinal thoracic aortic aneurysm (TAA) mesh datasets, we construct a new 3D dataset, named TAAMesh, using CT scans collected at Hammersmith Hospital, Imperial College London. The study was approved by the Health Research Authority (23/HRA/3733). The dataset consists of 590 CT scans from 208 patients diagnosed with thoracic aortic aneurysm, with each patient contributing 2 to 8 longitudinal scans. In addition to imaging data, the clinical information such as patient age, sex, maximum inscribed sphere (MIS) diameter, and corresponding clinical reports is also available. The distributions of patient age and MIS diameter with respect to sex are presented in Fig. \ref{fig:statistics}. 

\begin{figure*}[ht]
  \centering
   \includegraphics[width=1\linewidth]{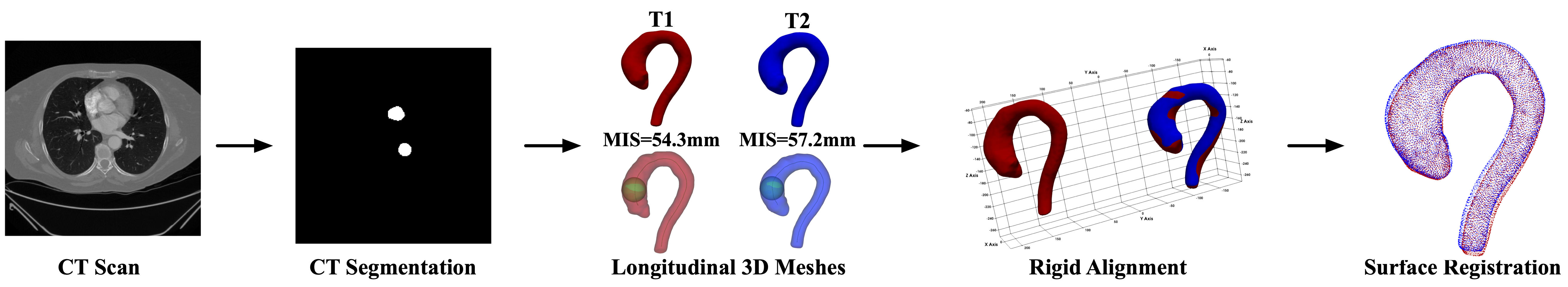}
   \caption{Overview of the CT-to-mesh preprocessing pipeline: (a) CT segmentation. The pre-trained TotalSegmentator model \cite{chung2024artificial} automatically delineates the aorta. Manual corrections are subsequently applied to eliminate any residual inaccuracies and guarantee pixel-level precision. (b) Mesh extraction. Binary segmentation masks are converted into triangular surface meshes with the marching-cubes algorithm implemented in the IRTK library. (c) The rigid alignment algorithm in IRTK brings every mesh into a common pose (same position and orientation). (d) The non-rigid surface registration algorithm in IRTK warps each mesh to establish one-to-one vertex correspondence.}
   \label{fig:preprocessing}
\end{figure*}
The CT-to-mesh preprocessing pipeline is illustrated in Fig. \ref{fig:preprocessing}. To convert CT scans into 3D mesh data, we first perform CT segmentation to obtain segmentation masks, utilizing the publicly available TotalSegmentator model \cite{wasserthal2023totalsegmentator}. After initial segmentation, manual corrections are applied to ensure pixel-level accuracy. Next, 3D meshes are generated from the segmentation masks using the marching cubes algorithm from the IRTK library \footnote{\url{https://www.doc.ic.ac.uk/~dr/software/usage.html}}. Rigid alignment is applied to align the mesh positions and orientations, followed by non-rigid surface registration for vertex correspondence matching, both provided by the IRTK library. Following these two registration steps, each mesh is standardized to 10,000 vertices. Finally, the mesh dataset is split into training, validation, and test sets at patient level, using a 7:1:2 ratio. 

We employ standard mesh reconstruction evaluation metrics (unit: mm), including Chamfer Distance (CD), Mean Absolute Error (MAE), and Hausdorff Distance (HD), as well as a clinically relevant metric--the MIS diameter error. The mean and standard deviation values are reported for CD, MAE and HD.

\subsection{Baselines and Implementation Details}

To the best of our knowledge, there are currently no prior works that have released source code for conditional mesh-to-mesh translations. Nonetheless, a variety of graph convolutional network (GCN) architectures have been developed for point cloud and mesh data analysis. From these, we adapt several widely used ones to construct baselines for comparison, including GCNConv \cite{kipf2017semi}, PointNet++ \cite{qi2017pointnet++}, GCN2Conv \cite{chen2020simple}, SSGConv \cite{zhu2021simple}, and PMLP \cite{yang2022graph}. To ensure fair comparisons, we build a deep model with four blocks for each type of GCN layer and incorporate the same condition branch for aneurysm growth prediction as the proposed method. All GCN layers used in the comparisons were implemented using the publicly available PyTorch Geometric library \cite{Fey/Lenssen/2019}. All the experiments are conducted on a server with an Intel(R) Xeon(R) Silver 4114 CPU @ 2.20GHz and a single Tesla V100 GPU with a 32GB memory. All models are trained for 1,000 epochs with a mini-batch size of 1 using the Adam optimization algorithm ($\alpha$ = 0.0002, $\beta_1$= 0.5). $K$ is set to 8 empirically.

\begin{figure*}[t]
  \centering
   \includegraphics[width=1.0\linewidth]{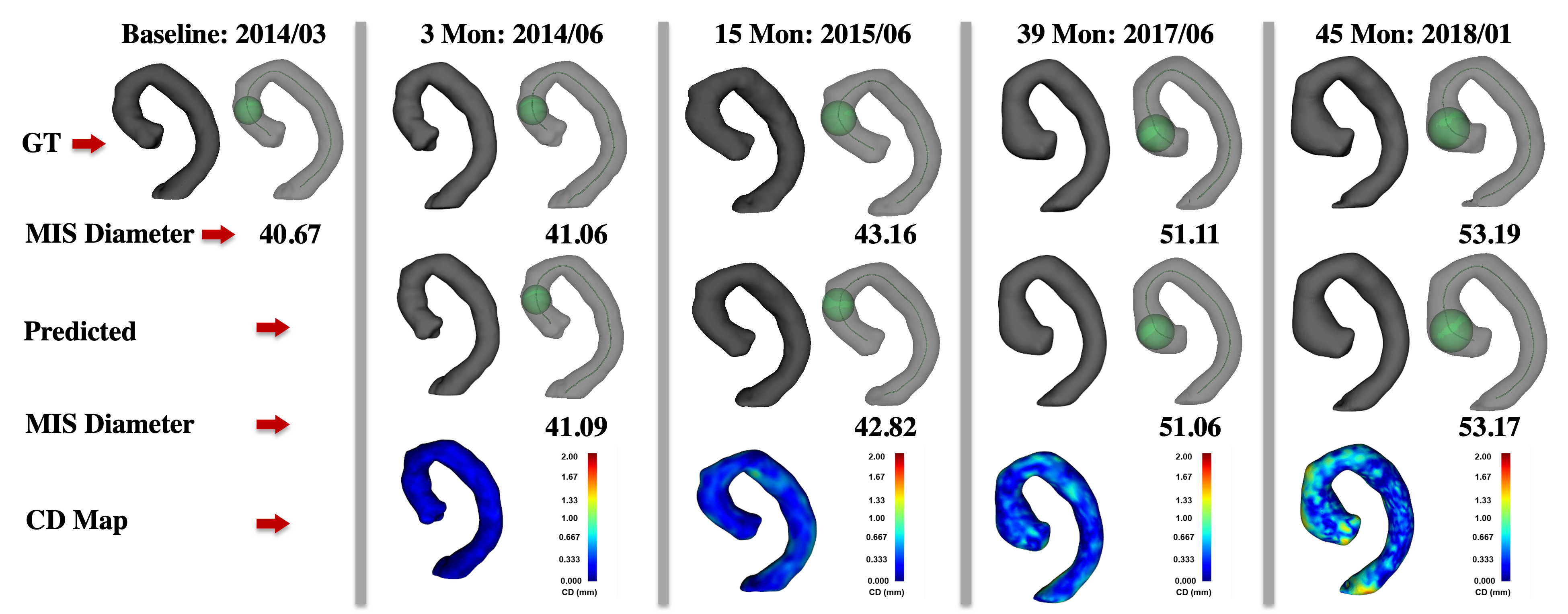}
   \caption{Given a ground truth (GT) baseline mesh (dated 2014/03), MCMeshGAN predicts follow-up mesh sequences based on the specified conditional time interval (in months). The color maps visualize the Chamfer Distance (CD) between the predicted meshes and the corresponding GT Follow-up meshes. The maximum diameter and centerline of the GT and predicted meshes are also provided for comparison.}
   \label{fig:demo}
\end{figure*}
\section{Results and Discussion}
Fig. \ref{fig:demo} illustrates MCMeshGAN’s ability to predict follow-up mesh sequences across different time intervals. The input (GT baseline) mesh to MCMeshGAN was collected in 2014/03, we set the time intervals to 3, 15, 39, and 45 months, respectively, to generate the corresponding future mesh sequences. The predicted meshes show strong agreement with the corresponding GT follow-up meshes, demonstrating the model’s ability to capture realistic anatomical changes over time. This is further supported by low Chamfer Distances and minimal errors in the clinically significant MIS diameter metric, highlighting MCMeshGAN’s accuracy and reliability for personalized, time-aware aneurysm growth prediction.

\begin{table*}\small
  \centering
  \setlength{\tabcolsep}{2.5mm}{
  \begin{tabular}{@{}l|l|cccc@{}}
    \toprule
    \textbf{Methods} & \textbf{\makecell[l]{Backbone Type \\(+Condition)}} & \textbf{MAE (mm)} & \textbf{CD (mm)} & \textbf{HD (mm)} & \textbf{MIS Error (mm)}\\ 
    \midrule
    PointNet++ \cite{qi2017pointnet++}   & DNN+Condition & 2.188$\pm$3.133 & 2.633$\pm$2.825 & 14.013$\pm$9.997 & 4.149\\
    GCNConv \cite{kipf2017semi}          & GCN+Condition & 2.042$\pm$3.363 & 2.740$\pm$2.302 & 13.878$\pm$9.571 & 4.154\\
    GCN2Conv \cite{chen2020simple}       & GCN+Condition & 1.976$\pm$3.460 & 2.670$\pm$2.412 & 13.547$\pm$9.673 & 3.977\\
    SSGConv \cite{zhu2021simple}         & GCN+Condition & 1.875$\pm$3.371 & 2.495$\pm$2.280 & 13.091$\pm$9.107 & 3.916\\
    PMLP \cite{yang2022graph}            & MLP+Condition & 1.952$\pm$2.946 & 2.651$\pm$2.005 & 13.372$\pm$9.758 &  4.059\\
    \midrule
    \midrule
    \textbf{Methods} & \textbf{\makecell[l]{Backbone Type \\(+Condition+KCN)}} & \textbf{MAE (mm)}& \textbf{CD (mm)} & \textbf{HD (mm)} & \textbf{MIS Error (mm)}\\
    \midrule
    PointNet++ \cite{qi2017pointnet++}   & DNN+Condition+KCN & 2.006$\pm$3.639 & 2.579$\pm$2.558 & 13.535$\pm$9.988 & 3.959\\
    GCNConv \cite{kipf2017semi}          & GCN+Condition+KCN & \makecell[c]{-} & - & - & -\\
    GCN2Conv \cite{chen2020simple}       & GCN+Condition+KCN & 1.605$\pm$3.516 & 2.228$\pm$2.457 & 12.161$\pm$9.534 & 3.385\\
    SSGConv \cite{zhu2021simple}         & GCN+Condition+KCN & 1.716$\pm$3.474 & 2.367$\pm$2.417 & 12.905$\pm$9.469 & 3.390\\
    PMLP \cite{yang2022graph}            & MLP+Condition+KCN & 1.743$\pm$2.821 & 2.486$\pm$1.927 & 13.055$\pm$8.330 & 3.873\\
    \midrule
    MCMeshGAN                            & GCN+Condition+KCN & \textbf{1.285$\pm$0.993} & \textbf{1.831$\pm$1.251} & \textbf{9.815$\pm$5.131} &  \textbf{2.887}\\
    \bottomrule
  \end{tabular}}
  \caption{Quantitative comparison of MCMeshGAN with baseline methods on the test set, in terms of mean absolute error (MAE) (unit: mm), Chamfer distance (CD) (unit: mm), Hausdorff Distance (HD) (unit: mm) and MIS error (unit: mm).}
  \label{tab:baselines}
\end{table*}

\begin{figure*}[t]
  \centering
   \includegraphics[width=1\linewidth]{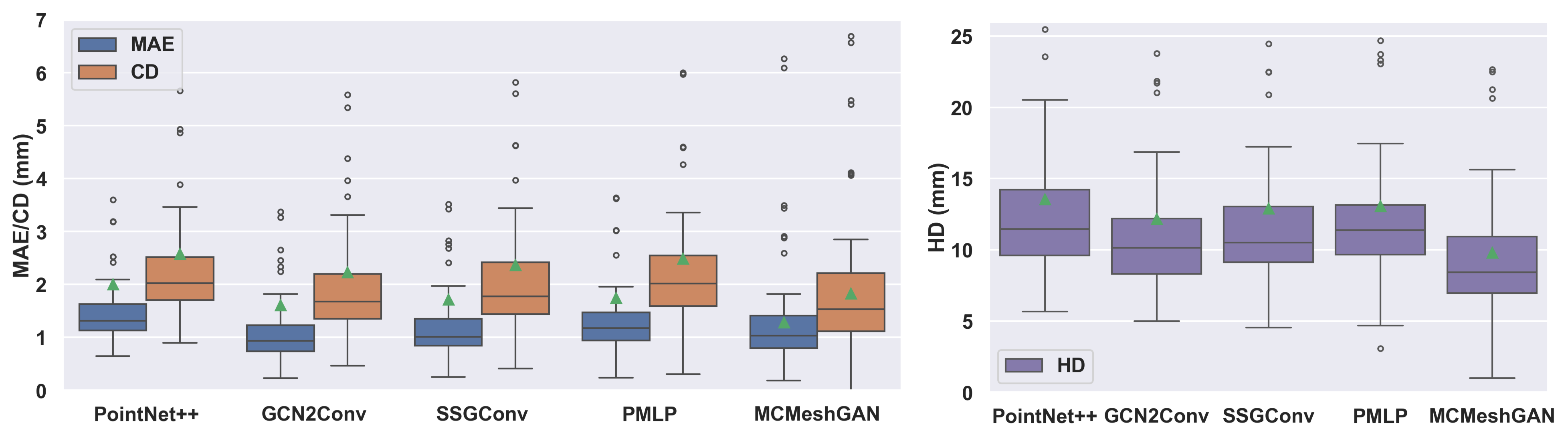}
   \caption{Performance comparisons of the evaluated methods. Box plots summarize three error metrics: mean average error (MAE), Chamfer distance (CD) and Hausdorff distance (HD). Each box shows the interquartile range (IQR); outliers appear as individual points, and green triangles indicate the mean.} 
   \label{fig:quantitative}
\end{figure*}

\subsection{Comparison with Baselines}
Table \ref{tab:baselines} and Fig. \ref{fig:quantitative} present the quantitative comparisons between the proposed MCMeshGAN model and various baseline models. The `+Condition' label indicates that our proposed condition branch has been integrated into the backbone of each baseline, as it is essential for incorporating condition inputs in aneurysm growth prediction. Furthermore, we introduce another set of experiments (denoted as `+Condition+KCN'), where our KCN branch is also added to the baselines. This is done to evaluate whether the KCN module can enhance the baselines by improving their ability to capture local geometric variations.

\begin{figure*}[t]
  \centering
   \includegraphics[width=1.0\linewidth]{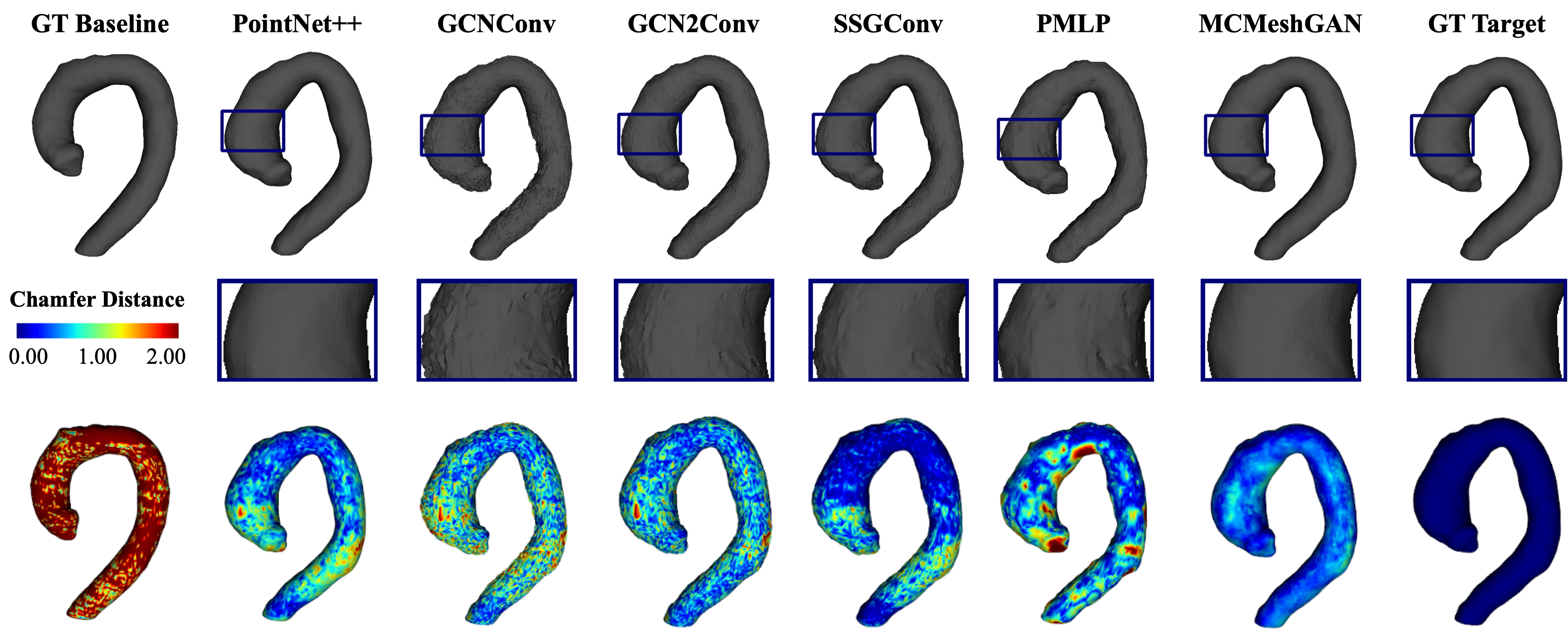}
   \caption{Qualitative comparison of MCMeshGAN with baseline frameworks. The color maps illustrate the Chamfer Distance between the predicted meshes and the GT target mesh at a 29-month interval.}
   \label{fig:baselines}
\end{figure*}

Table \ref{tab:baselines} presents the key results: (1) MCMeshGAN, which integrates KCN with GCN, achieves the lowest values in MAE, CD, and HD, indicating that its predicted meshes most closely resemble the ground truth in terms of geometry and structure. More importantly, MCMeshGAN also attains the lowest MIS diameter error, a clinically meaningful metric that reflects the accuracy of maximum diameter estimation--an essenstial factor in clinical guidelines for determining intervention thresholds. Unlike general geometric metrics, the MIS diameter is directly tied to clinical decision-making. Thus, superior performance on this metric suggests that MCMeshGAN is not only technically precise but also clinically reliable. (2) The KCN branch significantly enhances all GCN frameworks, particularly GCNConv and GCN2Conv. This is because KCN complements GCN by effectively capturing local details that GCNs may overlook. This is demonstrated by the qualitative comparisons shown in Fig. \ref{fig:baselines}. Without KCN, GCNConv produces a rough mesh surface that lacks many local details. In contrast, MCMeshGAN, which combines GCNConv and KCN, generates a smoother and more realistic mesh surface. (3) GCNConv obtains the best improvements from KCN among all the GCN frameworks. We believe this is because GCNConv captures global relationships more effectively by stacking multiple layer. However, as the network depth increases, it struggles to capture fine-grained local features without additional refinement. (4) SSGConv achieves the best performance among all the baselines without the support of KCN. This is because SSGConv focuses on spectral graph convolutions, which operate directly in the graph’s spectral domain. It excels at capturing local topological features through efficient spectral graph convolutions. 

Fig. \ref{fig:baselines} presents the qualitative comparison of MCMeshGAN with baseline frameworks. It is evident that nearly all GCN-based frameworks generate rough mesh surfaces in local regions. GCNConv and GCN2Conv produce the roughest surfaces, as they lose details in local regions after stacking multiple layers. In contrast, PointNet++ adopts a hierarchical structure to capture local features at multiple scales, while SSGConv leverages spectral graph convolutions to better capture geometric variations in local areas. Nevertheless, all the comparisons methods, excepted PointNet++, show some degree of surface roughness in the generated mesh. These observations suggest that introducing a hierarchical structure to captures local features at multiple scales could help address the issue of mesh surface roughness. Among all methods, our MCMeshGAN achieves the best Chamfer Distance map, as it uses KCN to capture local variations and GCN to preserve the overall geometry. However, slight surface roughness is still present in the results. 

\begin{table*}
  \centering
  \setlength{\tabcolsep}{2.3mm}{
  \begin{tabular}{@{}c|cc|cc|ccc|ccc@{}}
    \toprule
     \multirow{2}*{\textbf{Networks}} & \multicolumn{2}{|c|}{\textbf{Conditions}} & \multicolumn{2}{|c|}{\textbf{Backbones}} & \multicolumn{3}{|c|}{\textbf{Losses}} & \multicolumn{3}{c}{\textbf{Metrics}}\\
     \cmidrule{2-11}
     & \textbf{Age} & \textbf{sex} & \textbf{KCN} & \textbf{GCN} & \textbf{L$_1$} & \textbf{CD} & \textbf{Adv} & \textbf{MAE} & \textbf{CD} & \textbf{HD}\\
    \midrule
    \multirow{11}*{\rotatebox{90}{MCMeshGAN}}
    &   &   & \ding{52} & \ding{52} & \ding{52} &   & \ding{52}         & 1.533$\pm$2.348 & 2.125$\pm$2.066 & 11.078$\pm$9.112\\
    & \ding{52} &   & \ding{52} & \ding{52} & \ding{52} &   & \ding{52} & 1.394$\pm$1.535 & 1.984$\pm$1.796 & 10.534$\pm$7.947\\
    &   & \ding{52} & \ding{52} & \ding{52} & \ding{52} &   & \ding{52} & 1.503$\pm$1.769 & 2.097$\pm$2.084 & 10.739$\pm$8.569\\
    \cmidrule{2-11}
    & \ding{52} & \ding{52} & \ding{52} &   & \ding{52} &   & \ding{52} & 1.592$\pm$3.614 & 2.178$\pm$2.546 & 11.843$\pm$9.773\\
    & \ding{52} & \ding{52} &   & \ding{52} & \ding{52} &   & \ding{52} & 2.042$\pm$3.363 & 2.740$\pm$2.302 & 13.478$\pm$10.495\\
    & \ding{52} & \ding{52} & \ding{52} &   &   & \ding{52} & \ding{52} & 1.699$\pm$3.648 & 2.366$\pm$2.599 & 12.385$\pm$9.584\\
    & \ding{52} & \ding{52} &   & \ding{52} &   & \ding{52} & \ding{52} & 2.166$\pm$3.377 & 2.861$\pm$2.310 & 13.669$\pm$10.263\\   
    \cmidrule{2-11}
    & \ding{52} & \ding{52} & \ding{52} & \ding{52} & \ding{52} &   &   & 1.588$\pm$2.955 & 2.158$\pm$2.459 & 11.248$\pm$9.258\\
    & \ding{52} & \ding{52} & \ding{52} & \ding{52} &   & \ding{52} &   & 1.943$\pm$3.184 & 2.629$\pm$2.295 & 12.949$\pm$9.145\\
    & \ding{52} & \ding{52} & \ding{52} & \ding{52} &   & \ding{52} & \ding{52} & 1.515$\pm$2.100 & 2.357$\pm$1.627 & 11.349$\pm$9.339\\
    \cmidrule{2-11}
    & \ding{52} & \ding{52} & \ding{52} & \ding{52} & \ding{52} &   & \ding{52} & \textbf{1.285$\pm$0.993} & \textbf{1.831$\pm$1.251} & \textbf{9.815$\pm$5.131}\\
    \bottomrule
  \end{tabular}}
  \caption{Ablation experiment results on the test set, reported in terms of Mean Absolute Error (MAE), Chamfer Distance (CD) and Hausdorff Distance (HD) metrics.}
  \label{tab:ablation}
\end{table*}

\subsection{Ablation Study}
\normalcolor
To further verify the effect of each component in MCMeshGAN, we conduct the ablation study and present the results in Table \ref{tab:ablation}, several components are considered, including the age, sex, network backbones, and losses. Three losses are tested including L1, CD, and adversarial losses.

\subsubsection{Impact of Age and Sex}

From the results in Table \ref{tab:ablation}, we observe that age has a greater impact on model performance than sex. This can be attributed to the broader and more continuous variability present in the age feature vector, which spans a range from 35 to 93 years (as illustrated in Fig. \ref{fig:statistics}). In contrast, sex is a binary variable (male or female), offering limited granularity for the model to learn from. The wider dynamic range of the age feature provides the model with richer contextual cues about anatomical differences associated with aging. Consequently, the network becomes more sensitive to age-related morphological variations, and age emerges as a more influential factor in shaping the model’s predictions.

\begin{figure*}[ht]
  \centering
   \includegraphics[width=1\linewidth]{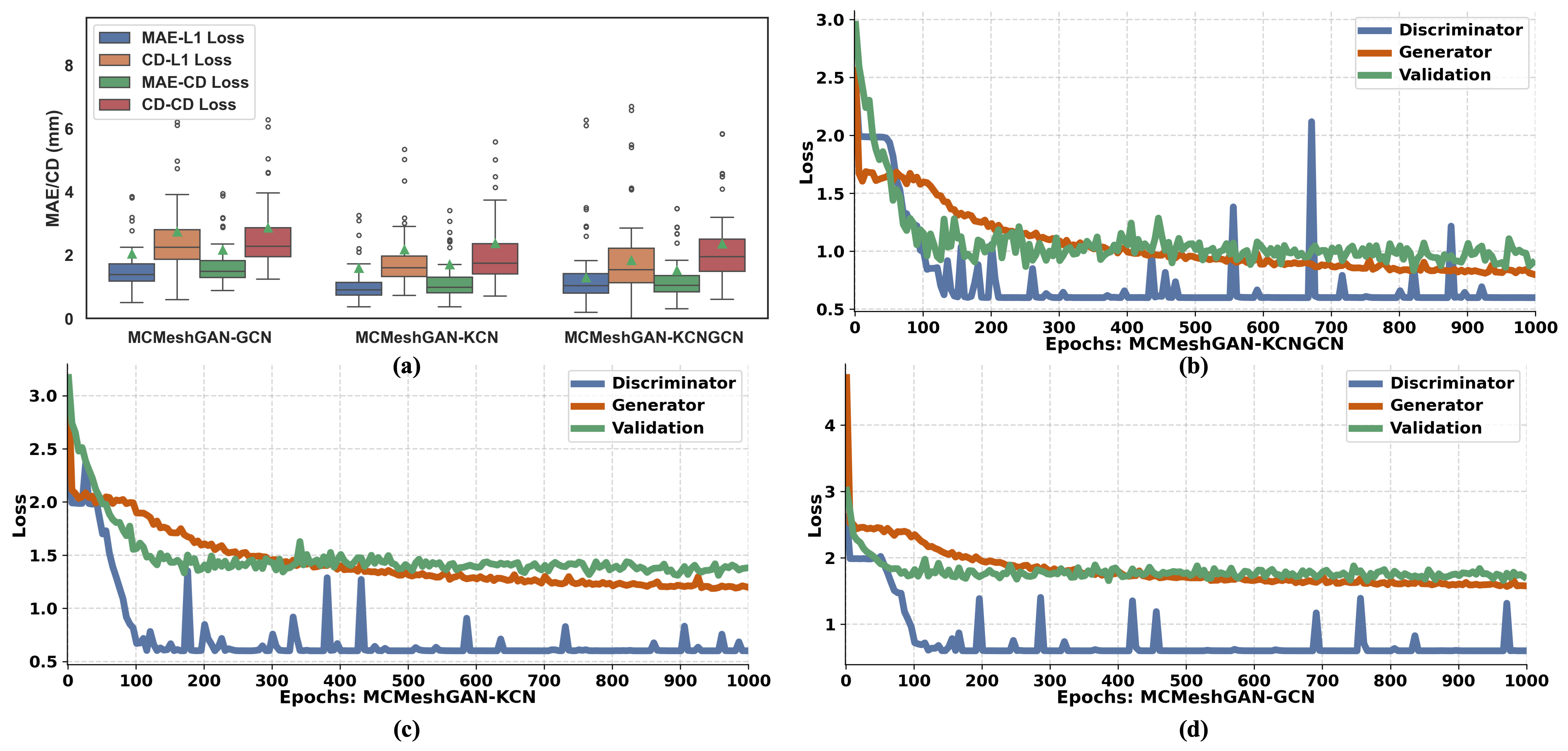}
   \caption{Performance comparisons (a) and losses (b, c, d) for MCMeshGANs with various backbones and losses. Plot (a) shows box plots of MCMeshGANs with different backbones and loss functions (L1 or CD loss): median and interquartile range (IQR) displayed, with mean value indicated by green triangle. Plots (b, c, d) display the training and validation losses for different MCMeshGANs using L1 loss.}
   \label{fig:loss_acc}
\end{figure*}
\begin{figure*}[t]
  \centering
   \includegraphics[width=1\linewidth]{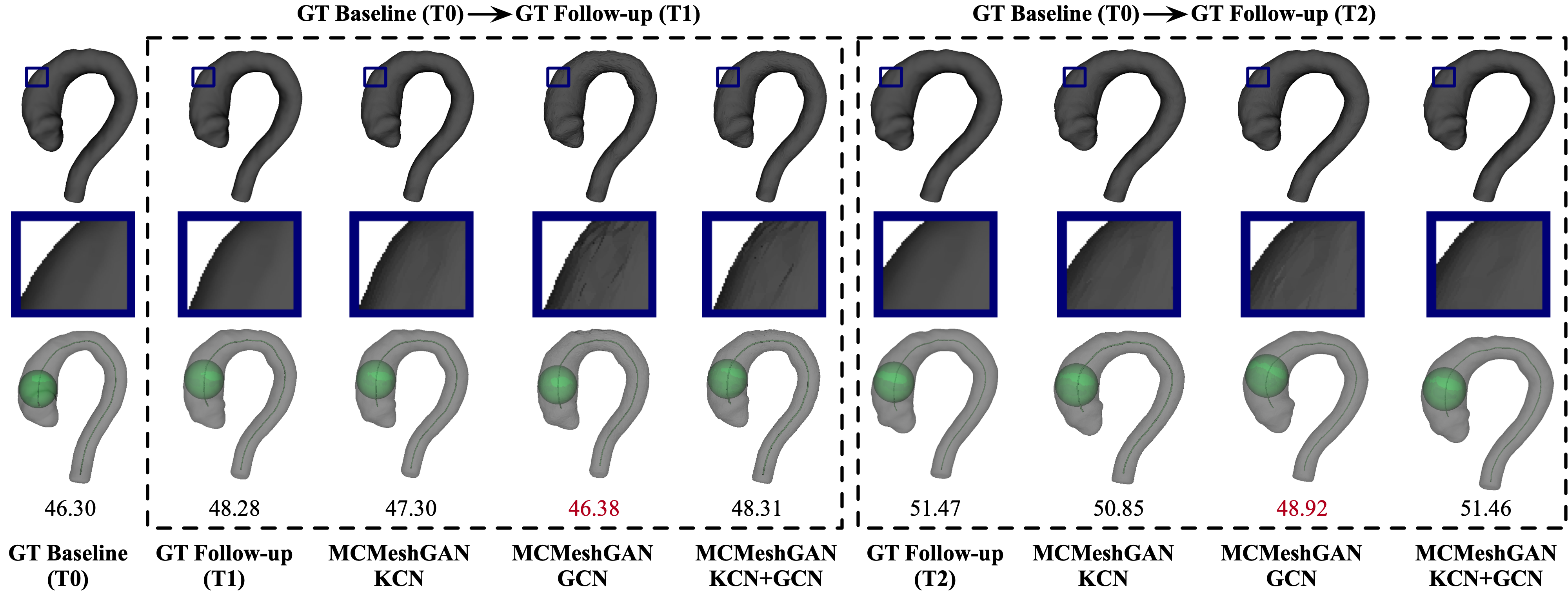}
   \caption{Qualitative results of MCMeshGAN with different backbones. The MIS diameter and center-line shown in the bottom row are computed using the open-source software 3D Slicer. MCMeshGAN-GCN produces the largest measurement errors.}
   \label{fig:backbone}
\end{figure*}
\subsubsection{Impact of Backbones} We design three backbones to evaluate their impact on MCMeshGAN: MCMeshGAN-KCNGCN with both KCN and GCN branches, MCMeshGAN-KCN with only the KCN branch, and MCMeshGAN-GCN with only the GCN branch. Fig. \ref{fig:loss_acc} presents performance and loss comparisons among three models, MeshGAN-KCNGCN consistently outperforms the other two, regardless of whether L1 or CD loss is used. Additionally, its generator and validation losses converges to lower values than those of the other models. These results indicate that KCN and GCN are complementary components that together enhance generation quality. Furthermore, MCMeshGAN-KCN achieves better MAE and CD scores than MCMeshGAN-GCN, likely because the GCN-only branch struggles to capture precise local variations, resulting in large vertex-wise differences. This is evident from the qualitative comparisons in Fig. \ref{fig:backbone}, where MCMeshGAN-GCN produces much rougher mesh surface than MCMeshGAN-KCN, leading to significant errors when measuring the maximum aneurysm diameter. In contrast, MCMeshGAN, which combines GCN with KCN, generates a smoother surface and achieve more accurate diameter measurements.

\subsubsection{Impact of Losses} We evaluate the contribution of three losses functions in shaping the performance of MCMeshGAN: L1 loss, CD loss, and adversarial loss. As shown in Table \ref{tab:ablation}, incorporating adversarial loss leads to noticeable improvements in quantitative metrics. This improvement is attributed to the adversarial loss's emphasis on perceptual realism, which encourages the generation of more anatomically plausible and visually natural mesh structures. Furthermore, L1 loss consistently outperforms CD loss, likely due to the non-rigid surface alignment algorithm used in the data preprocessing stage (Section IV-A), which tends to establish a one-to-one correspondence between vertices. L1 loss minimizes the absolute difference between corresponding vertices in the predicted and target meshes, preserving vertex structure and layout more effectively in one-to-one mapping. In contrast, CD loss, which measures nearest-neighbor distances, allows more flexibility in point matching but can result in slight misalignments.

\section{Conclusion and Further Work}
\normalcolor
In this paper, we proposed MCMeshGAN, a conditional mesh-to-mesh translation model for predicting 3D aneurysm growth and deformation. The model effectively captures both global contextual and local geometric features by integrating Graph Convolutional Networks (GCNs) and KNN-based Convolutional Networks (KCNs) in a dual-branch architecture. This design allows the model to simultaneously encode coarse anatomical structure and fine-grained surface details. To support further advancements, we present a discussion below on current limitations and potential steps for improvement:

\noindent\textbf{Rough Mesh Surface.} Generated meshes can still appear rough, which impairs visual quality and clinical accuracy. Future work could explore multi-scale models and loss functions that better preserve surface smoothness and mesh structure.

\noindent \textbf{Generalization to Broader Application Scenarios.} Our model was only tested on our private dataset, as no public longitudinal aneurysm mesh datasets exist. In the future, we plan to test on external data and collaborate with clinical partners to improve generalization.

\noindent \textbf{Clinical Validation and Integration.} The model has not yet been tested in real clinical workflows. Future steps include working with clinicians to validate results, compare with patient outcomes, and integrate the model into existing clinical workflow for aneurysm monitoring and treatment planning.

The proposed application of this technique will be to estimate future aneurysm evolution at arbitrary time intervals from the baseline diagnosis of TAA to develop individualized surveillance strategies and personalized risk prediction.  Current practice of annual or biannual imaging may not be cost-effective according a meta-analysis which urged further studies to inform the timing in which more frequent follow-up is required \cite{guo2018association}. 


\end{document}